%% file: main.tex
\documentclass[10pt,journal,compsoc]{IEEEtran}

\ifCLASSOPTIONcompsoc
  \usepackage[nocompress]{cite}
\else
  \usepackage{cite}
\fi

\ifCLASSINFOpdf
\else
\fi

\input{utils/packages}
\input{utils/macros}

\begin{document}
\title{LiDAR-based 4D Panoptic Segmentation \\ via Dynamic Shifting Network}

\author{Fangzhou~Hong, Hui~Zhou, Xinge~Zhu, Hongsheng~Li, Ziwei~Liu

\IEEEcompsocitemizethanks{
\IEEEcompsocthanksitem Fangzhou Hong and Ziwei Liu are with Nanyang Technological University. E-mail: \{fangzhou001, ziwei.liu\}@ntu.edu.sg.
\IEEEcompsocthanksitem Hui Zhou is with SenseTime Research. E-mail: smarthuizhou@gmail.com.
\IEEEcompsocthanksitem Xinge Zhu and Hongsheng Li are with the Chinese University of Hong Kong. E-mail: zx018@ie.cuhk.edu.hk, hsli@ee.cuhk.edu.hk.
}
}

\markboth{IEEE TRANSACTIONS ON PATTERN ANALYSIS AND MACHINE INTELLIGENCE, VOL. X, NO. X, MMMMMMM YYYY}%
{HONG \MakeLowercase{\textit{et al.}}: LiDAR-based Panoptic Segmentation via Dynamic Shifting Network}

\IEEEtitleabstractindextext{
\input{chaps/00_abstract}

}

\maketitle

\IEEEdisplaynontitleabstractindextext

\IEEEpeerreviewmaketitle

\IEEEraisesectionheading{\section{Introduction}\label{sec:intro}}
\input{chaps/pics/01_teaser}
\input{chaps/01_intro}

\section{Related Work}
\input{chaps/02_related_works}

\input{chaps/pics/03_dsnet_arch}
\section{Our Approach}
\input{chaps/03_00_overview}
\subsection{Strong Backbone Design}\label{3.1}
\input{chaps/03_01_baseline}
\subsection{Dynamic Shifting}\label{3.2}
\input{chaps/03_02_01_cluster}
\input{chaps/pics/03_04_4D-DS-Net-arch}
\input{chaps/03_02_02_dynamic_shift}
\subsection{Consensus-driven Fusion} \label{3.3}
\input{chaps/03_03_fusion}
\input{chaps/pics/04_02_ablation}
\input{chaps/03_04_4D}

\section{Experiments}
\input{chaps/tabs/04_02_tab_semkitti}
\input{chaps/04_01_setting}
\input{chaps/tabs/04_03_tab_nuscenes}

\input{chaps/04_02_semkitti}
\input{chaps/04_03_nuscenes}

\input{chaps/04_04_4D}

\input{chaps/04_05_analyze}

\section{Conclusion}
\input{chaps/05_conc}

\noindent \textbf{Acknowledgments} \quad
This research was conducted in collaboration with SenseTime. This work is supported by NTU NAP and A*STAR through the Industry Alignment Fund - Industry Collaboration Projects Grant. This work is supported in part by Centre for Perceptual and Interactive Intelligence Limited, in part by the General Research Fund through the Research Grants Council of Hong Kong under Grants (Nos. 14208417 and 14207319), in part by CUHK Strategic Fund.



\bibliographystyle{IEEEtran}
\bibliography{ref}

\clearpage

\input{utils/biography}

\end{document}

%% file: utils/packages.tex
\usepackage{epsfig}
\usepackage{graphicx}
\usepackage{amsmath}
\usepackage{amssymb}
\usepackage{caption}
\usepackage[ruled,vlined,linesnumbered]{algorithm2e}
\usepackage{enumitem}
\usepackage[misc]{ifsym}
\usepackage{makecell}
\usepackage{cases}
\usepackage[pagebackref=false,breaklinks=true,colorlinks=false,bookmarks=false]{hyperref}

%% file: utils/macros.tex
\newcommand{\nickname}{DS-Net}
\newcommand{\fullname}{Dynamic Shifting Network}
\newcommand{\things}{\textit{things}}
\newcommand{\stuff}{\textit{stuff}}
\newcommand{\PQ}{PQ}
\newcommand{\PQda}{PQ\textsuperscript{$\dagger$}}
\newcommand{\RQ}{RQ}
\newcommand{\SQ}{SQ}
\newcommand{\PQth}{PQ\textsuperscript{Th}}
\newcommand{\RQth}{RQ\textsuperscript{Th}}
\newcommand{\SQth}{SQ\textsuperscript{Th}}
\newcommand{\PQst}{PQ\textsuperscript{St}}
\newcommand{\RQst}{RQ\textsuperscript{St}}
\newcommand{\SQst}{SQ\textsuperscript{St}}
\newcommand{\miou}{mIoU}
\newcommand{\lstq}{LSTQ}
\newcommand{\scls}{S\textsubscript{cls}}
\newcommand{\sass}{S\textsubscript{assoc}}
\newcommand{\ioust}{IoU\textsuperscript{st}}
\newcommand{\iouth}{IoU\textsuperscript{th}}

\makeatletter
\DeclareRobustCommand\onedot{\futurelet\@let@token\@onedot}
\def\@onedot{\ifx\@let@token.\else.\null\fi\xspace}

\def\eg{\emph{e.g}\onedot} 
\def\ie{\emph{i.e}\onedot}

 \def\etal{\emph{et al}\onedot}
\makeatother

%% file: chaps/00_abstract.tex
\begin{abstract}
With the rapid advances of autonomous driving, it becomes critical to equip its sensing system with more holistic 3D perception.
However, existing works focus on parsing either the objects (\eg cars and pedestrians) or scenes (\eg trees and buildings) from the LiDAR sensor.
In this work, we address the task of \textbf{LiDAR-based panoptic segmentation}, which aims to parse both objects and scenes in a unified manner.
As one of the first endeavors towards this new challenging task, we propose the \fullname{} (\nickname{}), which serves as an effective panoptic segmentation framework in the point cloud realm.
In particular, \nickname{} has three appealing properties:
\textbf{1) Strong backbone design.} \nickname{} adopts the cylinder convolution that is specifically designed for LiDAR point clouds. 
The extracted features are shared by the semantic branch and the instance branch which operates in a bottom-up clustering style.
\textbf{2) Dynamic Shifting for complex point distributions.} We observe that commonly-used clustering algorithms like BFS or DBSCAN are incapable of
handling complex autonomous driving scenes with non-uniform point cloud distributions and varying
instance sizes.
Thus, we present an efficient learnable clustering module, dynamic shifting, which adapts kernel functions on the fly for different instances.
\textbf{3) Extension to 4D prediction.} Furthermore, we extend \nickname{} to 4D panoptic LiDAR segmentation by the temporally unified instance clustering on aligned LiDAR frames.
To comprehensively evaluate the performance of LiDAR-based panoptic segmentation, we construct and curate benchmarks from two large-scale autonomous driving LiDAR datasets, SemanticKITTI and nuScenes.
Extensive experiments demonstrate that our proposed \nickname{} achieves superior accuracies over current state-of-the-art methods in both tasks.
Notably, in the single frame version of the task, we outperform the SOTA method by 1.8\% in terms of the PQ metric. In the 4D version of the task, we surpass 2nd place by 5.4\% in terms of the LSTQ metric.
\end{abstract}

%% file: chaps/pics/01_teaser.tex

\begin{figure*}
    \begin{center}
        \centering
        \includegraphics[width=1.0\textwidth]{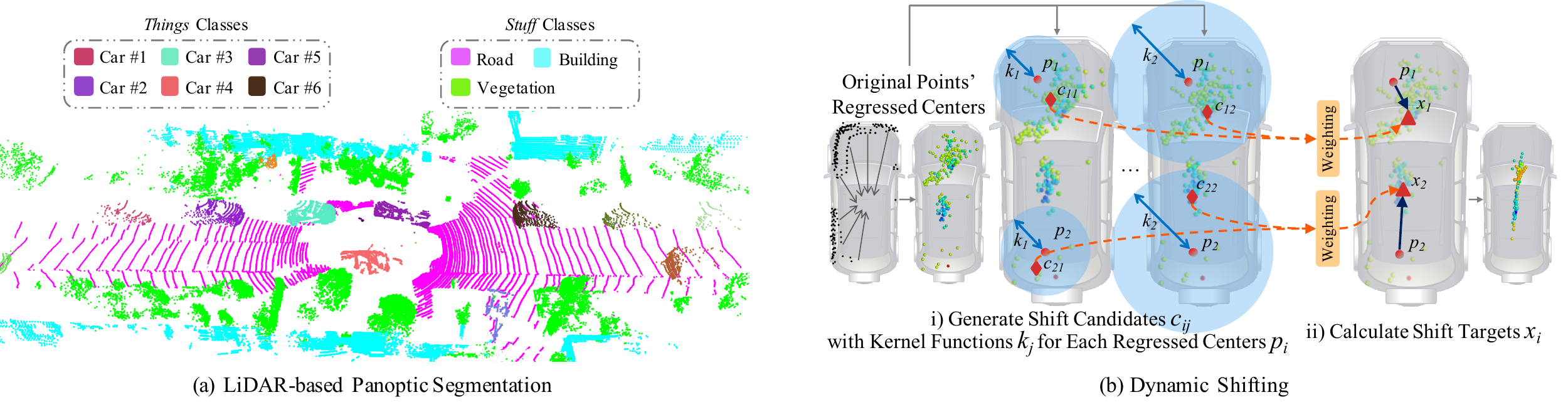}
        \captionof{figure}{As shown in (a), LiDAR-based panoptic segmentation requires instance-level segmentation for \textit{things} classes and semantic-level segmentation for \textit{stuff} classes. (b) shows the core operation of the proposed dynamic shifting where several shift candidates are weighted to obtain the optimal shift target for each regressed center.}
        \label{fig:01_teaser}
    \end{center}
\end{figure*}

%% file: chaps/01_intro.tex
\IEEEPARstart{A}{utonomous} driving, one of the most promising applications of computer vision, has achieved rapid progress in recent years.
Perception system, one of the most important modules in autonomous driving, has also attracted extensive studies in previous research works.
Admittedly, the classic tasks of 3D object detection \cite{lang2019pointpillars, shi2020pv, yan2018second} and semantic
segmentation \cite{milioto2019rangenet++, wu2018squeezeseg, zhang2020polarnet} have developed relatively mature solutions that support real-world autonomous driving prototypes.
However, there still exists a considerable gap between the existing works and the goal of holistic perception which is essential
for the challenging autonomous driving scenes.
In this work, we propose to close the gap by exploring the task of LiDAR-based panoptic segmentation, which requires full-spectrum point-level predictions.

Panoptic segmentation has been proposed in 2D detection \cite{kirillov2019panoptic} as a new vision task which unifies
semantic and instance segmentation.
Behley \etal \cite{behley2020benchmark} extend the task to LiDAR point clouds and propose the task of LiDAR-based panoptic segmentation.
As shown in Fig. \ref{fig:01_teaser} (a), this task requires to predict point-level semantic labels
for background (\stuff{}) classes (\eg road, building and vegetation),
while instance segmentation needs to be performed for foreground (\things{}) classes (\eg car, person and cyclist).

Nevertheless, the complex point distributions of LiDAR data make it difficult to perform reliable panoptic segmentation.
Most existing point cloud instance segmentation methods \cite{engelmann20203d, jiang2020pointgroup} are mainly designed for dense and uniform indoor point clouds.
Therefore, decent segmentation results can be achieved through the center regression and heuristic clustering algorithms.
However, due to the non-uniform density of LiDAR point clouds and varying sizes of instances,
the center regression fails to provide ideal point distributions for clustering.
The regressed centers usually form noisy strip distributions that vary in density and sizes.
As will be analyzed in Section \ref{3.2}, several heuristic clustering algorithms widely used in previous
works cannot provide satisfactory clustering results for the regressed centers of LiDAR point clouds.
To tackle the above mentioned technical challenges, we propose \fullname{} (\nickname{}) which is specifically designed for effective panoptic segmentation of LiDAR point clouds.

Firstly, we adopt a \textbf{strong backbone design} and provide a strong baseline for the new task.
Inspired by \cite{zhou2020cylinder3d}, the cylinder convolution is used
to efficiently extract grid-level features for each LiDAR frame in one pass which are further shared by the semantic and instance
branches.

Secondly, we present a novel \textbf{Dynamic Shifting Module} designed to cluster on the regressed centers with complex distributions
produced by the instance branch.
As illustrated in Fig. \ref{fig:01_teaser} (b), the proposed dynamic shifting module shifts the regressed centers to
the cluster centers.
The shift targets $x_i$ are adaptively computed by weighting across several shift candidates $c_{ij}$ which are
calculated through kernel functions $k_{j}$.
The special design of the module makes the \textit{shift} operation capable of dynamically adapting to the density or sizes of different
instances and therefore shows superior performance on LiDAR point clouds.
Further analysis also shows that the dynamic shifting module is robust and not sensitive to parameter settings.

Thirdly, the \textbf{Consensus-driven Fusion Module} is presented to unify the semantic and instance results
to obtain panoptic segmentation results.
The proposed consensus-driven fusion mainly solves the disagreement
caused by the class-agnostic style of instance segmentation.
The fusion module is highly efficient, thus brings negligible computation overhead.

With the single frame version of DS-Net established, we further extend it to the new task of 4D panoptic LiDAR segmentation, which is first introduced by \cite{aygun20214d}. The task not only requires panoptic segmentation for each frame, but also consistent things IDs across frames. In order to achieve that, we propose the temporally unified instance clustering on the aligned and overlapped consequent frames to achieve instance segmentation and instance association at the same time. Temporally consistent instance segmentation is further fused with the semantic segmentation to form the final 4D panoptic LiDAR segmentation results.

Extensive experiments on SemanticKITTI demonstrate the effectiveness of our proposed \nickname{}.
To further illustrate the generalizability of \nickname{}, we customize a LiDAR-based panoptic
segmentation dataset based on nuScenes.
As one of the first works for this new task, we present several strong baseline results by combining
the state-of-the-art semantic segmentation and detection methods.
\nickname{} outperforms all the state-of-the-art methods on both benchmarks (1st place on the public leaderboard of SemanticKITTI).

The main contributions are summarized below: 
\textbf{1)} To our best knowledge, we present one of the first attempts to address the challenging task of LiDAR-based panoptic segmentation.
\textbf{2)} The proposed \nickname{} effectively handles the complex distributions of LiDAR point clouds, and achieves state-of-the-art performance on SemanticKITTI and nuScenes.
\textbf{3)} We extend \nickname{} to the novel task of 4D panoptic LiDAR segmentation and achieve state-of-the-art results on SemanticKITTI.
\textbf{4)} Extensive experiments are performed on large-scale datasets. We adapt existing methods to this new task for in-depth comparisons. Further statistical analyses are carried out to provide valuable observations.

%% file: chaps/02_related_works.tex
\noindent\textbf{Single Frame Panoptic Segmentation.}
The challenging vision task of panoptic segmentation is firstly defined by \cite{kirillov2019panoptic} where semantic segmentation for
\stuff{} classes \cite{liu2015semantic} and instance segmentation for \things{} classes are evaluated under unified metrices.
From the perspective of network architecture, most panoptic segmentation methods can be categorized into top-down style and bottom-up style.
The top-down methods are mostly based on MaskRCNN \cite{he2017mask} where the instances are firstly detected then segmented by predicting masks.
Main innovations of this kinds of methods lie in the following two aspects.
The first one \cite{kirillov2019panoptic2, porzi2019seamless, li2019attention, wu2020bidirectional} is the backbone where semantic and instance information are extracted and shared.
Panoptic FPN \cite{kirillov2019panoptic2} and Seamless \cite{porzi2019seamless} manage to share MaskRCNN Feature Pyramid Network (FPN) between semantic and instance branches which yields a solid and strong baseline for the emerging task.
The second aspect \cite{kirillov2019panoptic2, liu2019end, yang2019sognet, chen2020banet} is the handling of disagreement between semantic and instance segmentation predictions and conflicts between multiple instance segmentation predictions.
UPSNet \cite{xiong2019upsnet} and Li \etal \cite{li2020unifying} try to unify \things{} and \stuff{} segmentation by introducing panoptic logits which can generate coherent panoptic segmentation results without any post processing.

The bottom-up approaches \cite{wang2020pixel, cheng2020panoptic, wang2020axial} typically perform semantic segmentation first, then perform pixel clustering
based on the semantic predictions, which natually saves the trouble of conflict handling and would lead to
lighter network design.
Although top-down approaches tend to outperform bottom-up approaches due to the use of the powerful
MaskRCNN, recent work of Panoptic-DeepLab \cite{cheng2020panoptic} presents a bottom-up baseline that has comparable performance
with top-down methods.
For the simplicity of the network design, we choose to use bottom-up approach in the proposed \nickname{}.

\noindent\textbf{Video Panoptic Segmentation.}
With the development of single frame panoptic segmentation techniques, recent researches have extended the task to video inputs. \cite{kim2020video} first formally defined the task and evaluation metrics for video panoptic segmentation, which requires the consistency of things IDs across frames. \cite{kim2020video} constructs the method based on UPSNet \cite{xiong2019upsnet}. Consequent two frames are fused using spatial-temporal attention. Finally, an object-level tracking is performed for consistent things IDs. \cite{qiao2021vip}, however, adopts a bottom-up backbone. Center regression is performed for two consequent frames with the centers predicted from the first frame, which would naturally produce consistent things IDs.

\noindent\textbf{Point Cloud Semantic and Instance Segmentation.}
According to the data representations of point clouds, most point cloud semantic segmentation methods can be categorized to point-based and voxel-based methods.
Based on PointNet-like backbones \cite{qi2017pointnet, qi2017pointnet++, liu2020lrc}, KPConv \cite{thomas2019kpconv}, DGCNN \cite{wang2019dynamic}, PointConv \cite{wu2019pointconv} and Randla-Net \cite{hu2020randla} can directly operate on unordered point clouds.
However, due to space and time complexity, most point-based methods struggle on large-scale point clouds datasets \eg ScanNet \cite{dai2017scannet}, S3DIS \cite{armeni20163d}, and SemanticKITTI \cite{behley2019semantickitti}.
MinkowskiNet \cite{choy20194d} utilizes the sparse convolutions to efficiently perform semantic segmentation on the voxelized large-scale point clouds.
SqueezeSeg \cite{wu2018squeezeseg} views LiDAR point clouds as range images while PolarNet \cite{zhang2020polarnet} and Cylinder3D \cite{zhou2020cylinder3d,zhu2021cylindricalseg,zhu2021cylindricalper} divide the LiDAR point clouds under the polar and cylindrical coordinate systems.

Previous works have shown great progress in the instance segmentation of indoor point clouds.
A large number of point-based methods (\eg SGPN \cite{wang2018sgpn}, ASIS \cite{wang2019associatively}, JSIS3D \cite{pham2019jsis3d} and JSNet \cite{zhao2020jsnet}) split the whole scene into small blocks and learn point-wise embeddings for final clustering, which are limited by the heuristic post processing steps and the lack of perception.
To avoid the problems, recent works (\eg PointGroup \cite{jiang2020pointgroup}, 3D-MPA \cite{engelmann20203d}, OccuSeg \cite{han2020occuseg}) use sparse convolutions to extract features of the whole scene in one pass.
As for LiDAR point clouds, there are a few previous works \cite{hu2020randla, Wong2019IdentifyingUI, wu2018squeezeseg, 2020LiDARSeg} trying to tackle the problem.

\noindent \textbf{Point Cloud Panoptic Segmentation.}
Recently, several attempts \cite{behley2020benchmark, milioto2020iros, zhou2021panoptic, hong2021lidar} have been made in the emerging task of point cloud panoptic segmentation. \cite{behley2020benchmark} first formally defines the task of LiDAR-based panoptic segmentation and proposes to combine the semantic segmentation and 3D object detection to obtain the panoptic segmentation results. \cite{milioto2020iros} performs spherical projection on LiDAR point clouds and utilize a bottom-up 2D panoptic segmentation. \cite{zhou2021panoptic} utilizes the Polar BEV encoder \cite{zhang2020polarnet} to extract per-point features. A semantic segmentation and class-agnostic instance clustering technique is then used to obtain final results. \cite{aygun20214d} extend the single frame version of LiDAR-based panoptic segmentation to the 4D version by greedy cross-volume association based on the overlap score. Our proposed 4D panoptic LiDAR segmentation methods performs unified instance clustering on the overlapping consequent LiDAR scans to obtain consistent things IDs.

%% file: chaps/pics/03_dsnet_arch.tex
\begin{figure*}[t]
    \begin{center}
        \includegraphics[width=1.0\linewidth]{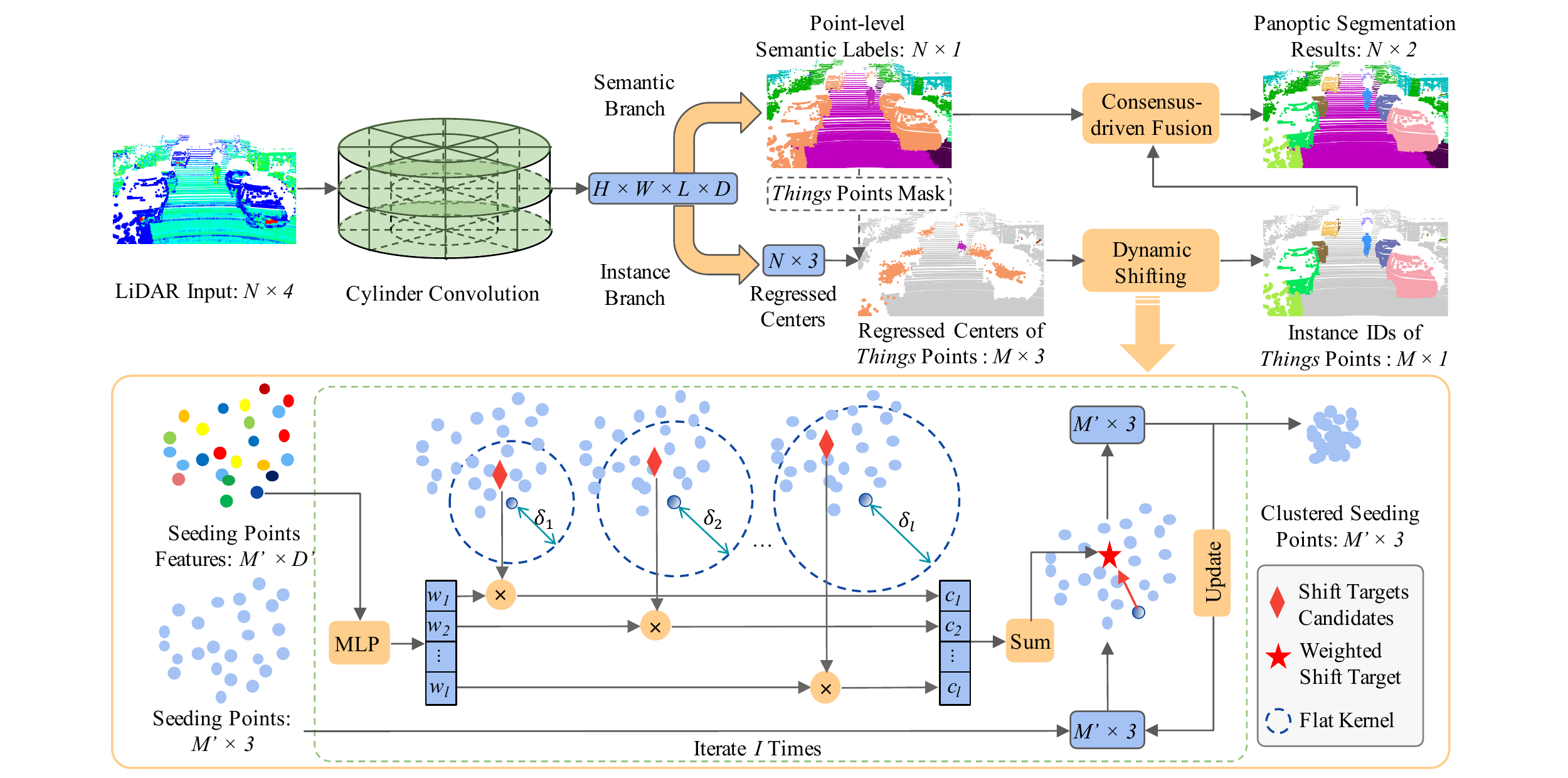}
    \end{center}
    \caption{\textbf{Architecture of the \nickname{}.} The \nickname{} consists of the cylinder convolution, a semantic and an instance branch as shown in the upper part of the figure. The regressed centers provided by the instance branch are clustered by the novel dynamic shifting module which is shown in the bottom half. The consensus-driven fusion module unifies the semantic and instance results into the final panoptic segmentation results.}
    \label{fig:03_dsnet_arch}
\end{figure*}

%% file: chaps/03_00_overview.tex
As one of the first attempts on the task of LiDAR-based panoptic segmentation, we first introduce a strong backbone
to establish a simple baseline (Sec. \ref{3.1}), based on which two modules are further proposed.
The novel dynamic shifting module is presented to tackle the challenge of the non-uniform LiDAR point clouds distributions (Sec. \ref{3.2}).
The efficient consensus-driven fusion module combines the semantic and instance predictions and produces panoptic segmentation results (Sec. \ref{3.3}).
The whole pipeline of the \nickname{} is illustrated in Fig. \ref{fig:03_dsnet_arch}.
Finally, we introduce a simple yet effective extension to the task of 4D panoptic LiDAR segmentation. (Sec. \ref{3.4})

%% file: chaps/03_01_baseline.tex
To obtain panoptic segmentation results, it is natural to solve two sub-tasks separately, which are semantic and instance segmentation, and combine the
results.
As shown in the upper part of Fig. \ref{fig:03_dsnet_arch}, the strong backbone consists of three parts:
the cylinder convolution, a semantic branch, and an instance branch.
High quality grid-level features are extracted by the cylinder
convolution from raw LiDAR point clouds and then shared by semantic and instance branches.

\noindent\textbf{Cylinder Convolution.}
Considering the difficulty presented by the task, we find that the cylinder convolution \cite{zhou2020cylinder3d} best
meets the strict requirements of high efficiency, high performance and fully mining of 3D positional relationship.
The cylindrical voxel partition can produce more even point distribution than normal Cartesian voxel
partition and therefore leads to higher feature extraction efficiency and higher performance.
Cylindrical voxel representation combined with sparse convolutions can naturally retain and fully explore
3D positional relationship.
Thus we choose the cylinder convolution as our feature extractor.

\noindent\textbf{Semantic Branch.}
The semantic branch performs semantic segmentation by connecting MLP to the cylinder convolution to predict semantic confidences for each voxel grid.
Specifically, the input LiDAR point clouds $P = \{p_i\} \in \mathbb{R}^{N\times 4}$, where $i \in \{1,...,N\}$, consists of
$N$ points and each point has four attributes $p_i = (x_i, y_i, z_i, r_i)$ representing its XYZ coordinates and
the intensity of the corresponding reflection beams.
The output of the backbone is the voxel features $F_v \in \mathbb{R}^{H\times W\times L\times D}$, where $C$
represents the dimension of the features.
As for semantic segmentation branch, by applying convolution to $F_v$, semantic logits
$L_s \in \mathbb{R}^{H\times W\times L\times C}$, where $C$ is the number of all classes,
are predicted for each voxel, which is then followed by max operation to compute the predicted semantic label
for each voxel.
Point-level semantic predictions are obtained by copying voxel labels to the points inside the voxels.
Considering the category imbalance in autonomous driving scene, we choose Weighted Cross Entropy and
Lovasz Loss as the loss function for semantic segmentation branch.

\noindent\textbf{Instance Branch.}
The instance branch utilizes center regression to prepare the \things{} points for further clustering.
The center regression module uses MLP to adapt cylinder convolution features and make \things{}
points to regress the centers of their instances by predicting the offset vectors $O \in \mathbb{R}^{M\times 3}$
pointing from the points $P \in \mathbb{R}^{M \times 3}$ to the instance centers $C_{gt} \in \mathbb{R}^{M\times 3}$.
The loss function for instance branch can be formulated as:
\begin{equation}
    L_{ins} = \frac{1}{M}\sum_{i=0}^{M}\lVert O[i] - (C_{gt}[i] - P[i]) \rVert_1 \text{,}
\end{equation}
where $M$ is the number of \things{} points.
The regressed centers $O + P$ are further clustered to obtain the instance IDs, which can be achieved
by either heuristic clustering algorithms or the proposed dynamic shifting module which are further introduced and
analyzed in the following section.

%% file: chaps/03_02_01_cluster.tex
\noindent\textbf{Point Clustering Revisit.}
Unlike indoor point clouds which are carefully reconstructed using RGB-D videos, the LiDAR point
clouds have the distributions that are not suitable for normal clustering solutions used by indoor instance segmentation methods.
The varying instance sizes, the sparsity and incompleteness of LiDAR point clouds make it difficult for the center
regression module to predict the precise center location and would result in noisy long strips distribution as displayed in
Fig. \ref{fig:01_teaser} (b) instead of an ideal ball-shaped cluster around the center.
Moreover, as presented in Fig. \ref{fig:03_02_density_and_msbandwidth} (a), the clusters formed by regressed centers that are far from the LiDAR sensor
have much lower densities than those of nearby clusters due to the non-consistent sparsity of LiDAR point clouds.
Facing the non-uniform distribution of regressed centers, heuristic clustering algorithms struggle to produce satisfactory results.
Four major heuristic clustering algorithms that are used in previous bottom-up indoor point clouds instance segmentation methods
are analyzed below.

\input{chaps/pics/03_02_density_and_msbandwidth}

\begin{itemize}[leftmargin=*]
    \item \textbf{Breadth First Search (BFS).}
        BFS is simple and good enough for indoor point clouds as proved in \cite{jiang2020pointgroup}, but not suitable
        for LiDAR point clouds.
        As discussed above, large density difference between clusters means that the \emph{fixed radius}
        cannot properly adapt to different clusters.
        Small radius will over-segment distant instances while large radius will under-segment near instances.
    \item \textbf{DBSCAN \cite{ester1996density} and HDBSCAN \cite{campello2013density}.}
        As density-based clustering algorithms, there is no surprise that these two algorithms also perform badly on the
        LiDAR point clouds, even though they are proved to be effective for clustering indoor point clouds
        \cite{engelmann20203d, zhang2020spatial}.
        The core operation of DBSCAN is the same as that of BFS.
        While HDBSCAN intuitively assumes that the points with lower density are more likely to be noise points which is
        not the case in LiDAR points.
    \item \textbf{Mean Shift \cite{comaniciu2002mean}.}
        The advantage of Mean Shift, which is used by \cite{lahoud20193d} to cluster indoor point clouds, is that the
        kernel function is not sensitive to density changes and robust to noise points which makes it more suitable than
        density-based algorithms.
        However, the \textit{bandwidth} of the kernel function has great impact on the clustering results as shown in Fig.
        \ref{fig:03_02_density_and_msbandwidth} (b).
        The fixed bandwidth cannot handle the situation of large and small instances simultaneously which makes Mean Shift
        also not the ideal choice for this task.
\end{itemize}

%% file: chaps/pics/03_02_density_and_msbandwidth.tex
\begin{figure}[t]
    \begin{center}
        \includegraphics[width=1.0\linewidth]{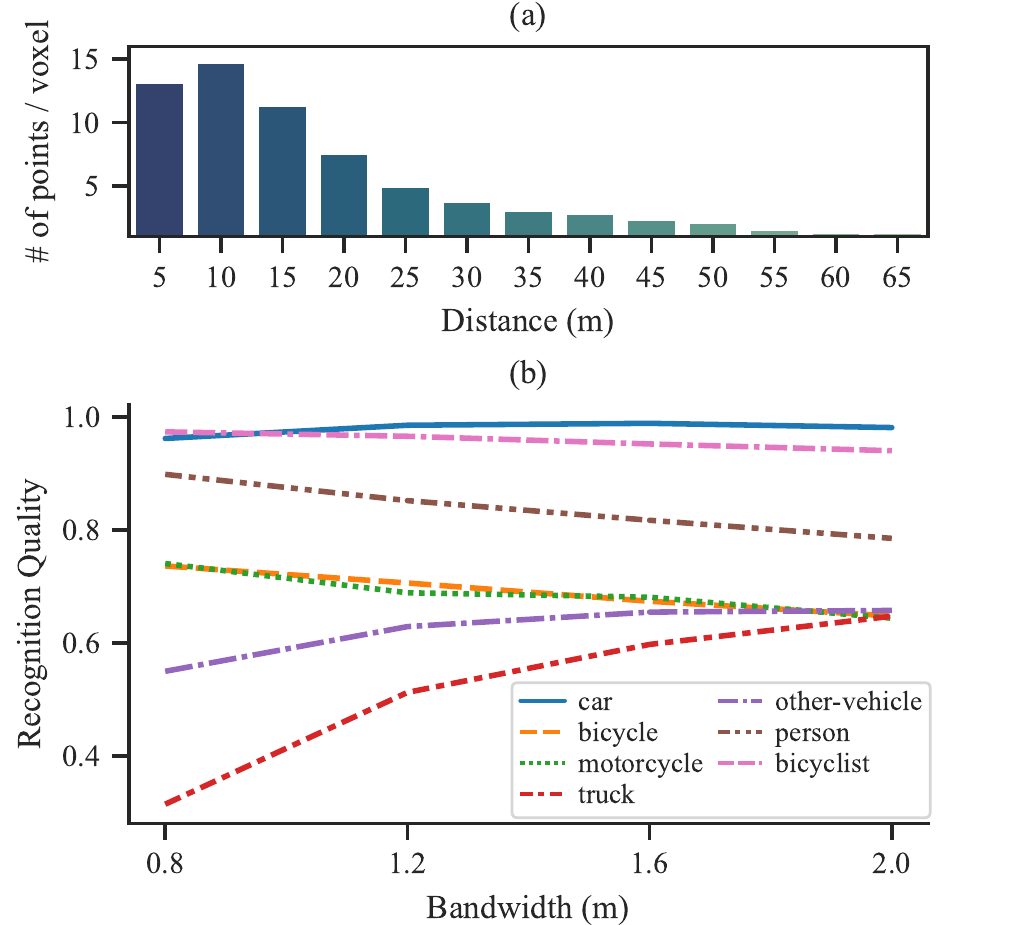}
    \end{center}
    \caption{(a) counts the average number of regressed centers inside each valid voxel of instances at different distances. (b) shows the effect of Different Mean Shift Bandwidth on the Recognition Quality of Different Classes.}
    \label{fig:03_02_density_and_msbandwidth}
\end{figure}

%% file: chaps/pics/03_04_4D-DS-Net-arch.tex
\begin{figure*}[ht]
    \begin{center}
        \includegraphics[width=1.0\linewidth]{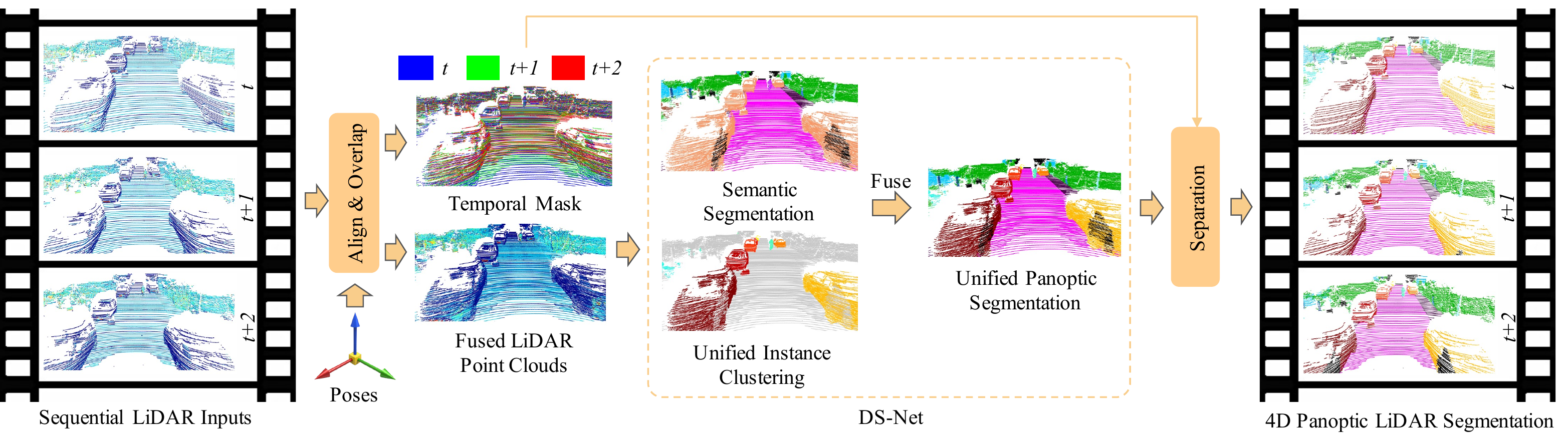}
    \end{center}
    \caption{\textbf{Architecture of the 4D version \nickname{} (namely 4D-\nickname{}).} The 4D-\nickname{} takes the aligned and overlapped LiDAR scans as input. Then perform unified instance clustering to generate temporally consistent things IDs. Finally the overlapped panoptic segmentation results are separated according to the original temporal masks.}
    \label{fig:03_04_4d}
\end{figure*}

%% file: chaps/03_02_02_dynamic_shift.tex
\input{chaps/algo/pseudocode}

\noindent\textbf{Dynamic Shifting.}
As discussed above, it is a robust way of estimating cluster centers of regressed centers by iteratively
applying kernel functions as in Mean Shift.
However, the fixed bandwidth of kernel functions fails to adapt to varying instance sizes.
Therefore, we propose the dynamic shifting module which can automatically adapt the kernel function for
each LiDAR point in the complex autonomous driving scene so that the regressed centers can be dynamically,
efficiently and precisely shifted to the correct cluster centers.

In order to make the kernel function learnable, we first consider how to mathematically define a differentiable
\textit{shift} opration.
Inspired by \cite{kong2018recurrent}, the shift operation on the seeding points (\ie points to be clustered)
can be expressed as matrix operations if the number of iterations is fixed.
Specifically, one iteration of shift operation can be formulated as follows.
Denoting $X\in \mathbb{R}^{M\times 3}$ as the $M$ seeding points, $X$ will be updated once
by the shift vector $S \in \mathbb{R}^{M\times 3}$ which is formulated as
\begin{equation}
    X \leftarrow X + \eta S \text{,}
\end{equation}
where $\eta$ is a scaling factor which is set to $1$ in our experiments.

The calculation of the shift vector $S$ is by applying kernel function $f$ on $X$, and formally defined as
$S = f(X) - X$.

Among various kinds of kernel functions, the flat kernel is simple but effective for generating shift target
estimations for LiDAR points, which is introduced as follows.
The process of applying flat kernel can be thought of as placing a query ball of certain radius (\ie bandwidth) centered at each seeding
point and the result of the flat kernel is the mass of the points inside the query ball.
Mathematically, the flat kernel $f(X)=D^{-1} K X$ is defined by the
kernel matrix $K=(X X^T \leq \delta)$, which masks out the points within a certain bandwidth $\delta$ for each seeding point,
and the diagonal matrix $D=diag(K \mathbf{1})$ that represents the number of points within the seeding point's bandwidth.

With a differentiable version of the shift operation defined, we proceed to our goal
of dynamic shifting by adapting the kernel function for each point.
In order to make the kernel function adaptable for instances with different sizes, the optimal bandwidth
for each seeding point has to be inferenced dynamically.
A natural solution is to directly regress bandwidth for each seeding point, which however is not differentiable
if used with the flat kernel.
Even though Gaussian kernel can make direct bandwidth regression trainable, it is still not the best solution
as analyzed in section \ref{ablation}.
Therefore, we apply the design of weighting across several bandwidth candidates to dynamically adapt to the optimal one.

One iteration of dynamic shifting is formally defined as follows.
As shown in the bottom half of Fig. \ref{fig:03_dsnet_arch}, $l$ bandwidth candidates
$L=\{\delta_1, \delta_2, ...,\delta_l\}$ are set.
For each seeding point, $l$ shift target candidates are calculated by $l$ flat kernels with corresponding
bandwidth candidates.
Seeding points then dynamically decide the final shift targets, which are ideally the closest to the cluster centers,
by learning the weights $W \in \mathbb{R}^{M\times l}$ to weight on $l$ candidate targets.
The weights $W$ are learned by applying MLP and Softmax on the backbone features so that
$\sum_{j=1}^{l}W[:, j] = \textbf{1}$.
The above procedure and the new learnable kernel function $\hat{f}$ can be formulated as
\begin{equation}
    \hat{f}(X) = \sum_{j=1}^{l} W[:, j] \odot (D_j^{-1} K_j X) \text{,}
\end{equation}
where $K_j = (X X^T \leq \delta_j)$ and $D_j = diag(K_j \mathbf{1})$.

With the one iteration of dynamic shifting stated clearly, the full pipeline of the dynamic shifting module, which is
formally defined in algorithm \ref{algo:forward_pass}, can be illustrated as follows.
Firstly, to maintain the efficiency of the algorithm, farthest point sampling (FPS) is performed on $M$ \things{}
points to provide $M'$ seeding points for the dynamic shifting iterations (Lines \ref{algo:1}--\ref{algo:2}).
After a fixed number $I$ of dynamic shifting iterations (Lines \ref{algo:3}--\ref{algo:12}), all seeding points have converged to the cluster centers.
A simple heuristic clustering algorithm is performed to cluster the converged seeding points to obtain
instance IDs for each seeding point (Line \ref{algo:13}).
Finally, all other \things{} points find the nearest seeding points and the corresponding instance IDs are
assigned to them (Lines \ref{algo:14}--\ref{algo:15}).

The optimization of dynamic shifting module is not intuitive since it is impractical to obtain the ground
truth bandwidth for each seeding point.
The loss function has to encourage seeding points shifting towards their cluster
centers which have no ground truths but can be approximated by the ground truth centers of instances
$C_{gt}' \in \mathbb{R}^{M' \times 3}$.
Therefore, the loss function for the $i$th iteration of dynamic shifting is defined by the manhattan distance between
the ground truth centers $C_{gt}'$ and the $i$th dynamically calculated shift targets $X_{i}$, which can be
formulated as
\begin{equation}
    l_{i} = \frac{1}{M'} \sum_{x=1}^{M'}\lVert X_{i}[x] - C_{gt}'[x] \rVert_1\text{.}
\end{equation}

Adding up all the losses of $I$ iterations gives us the loss function $L_{ds}$ for the dynamic shifting module:
    $L_{ds} = \sum_{i=1}^{I} w_{i} l_{i}\text{,}$
where $w_i$ are weights for losses of different iterations and are all set to $1$ in our experiments.

%% file: chaps/algo/pseudocode.tex
\begin{algorithm}[t]
    \SetAlgoLined
    \KwIn{\textit{Things} Points $P \in \mathbb{R}^{M\times 3}$, \textit{Things} Features $F\in\mathbb{R}^{M\times D'}$, \textit{Things} Regressed Centers $C\in\mathbb{R}^{M\times 3}$, Fixed number of iteration $I\in\mathbb{N}$, Bandwidth candidates list $L\in\mathbb{R}^{l}$}
    \KwOut{Instance IDs of \textit{things} points $R\in\mathbb{R}^{M\times 1}$}
    $mask = FPS(P)$, $P' = P[mask]$ \label{algo:1} \\
    $X = C[mask]$, $F' = F[mask]$ \label{algo:2} \\
    \For{$i \gets 1\ \KwTo\ I$} { \label{algo:3}
        $W_i = Softmax(MLP(F'))$\\
        $acc = zeros\_like(X)$\\
        \For{$j \gets 1\ \KwTo\ l$} {
            $K_{ij} = (X X^T \leq L[j])$\\
            $D_{ij} = diag(K_{ij} \mathbf{1})$\\
            $acc = acc + W_i[:, j] \odot (D_{ij}^{-1}K_{ij}X)$
        }
        $X = acc$\\
    } \label{algo:12}
    $R' = cluster(X)$ \label{algo:13} \\
    $index = nearest\_neighbour(P, P')$ \label{algo:14} \\
    $R = R'[index]$ \label{algo:15} \\
    \Return $R$
    \caption{Forward Pass of the Dynamic Shifting Module}
    \label{algo:forward_pass}
\end{algorithm}
\setlength{\textfloatsep}{6pt}

%% file: chaps/03_03_fusion.tex
Typically, solving the conflict between semantic and instance predictions is one of the essential steps in panoptic segmentation.
The advantages of bottom-up methods are that all points with predicted instance IDs must be in \things{} classes and one point will not be assigned to two instances.
The only conflict needs to be solved is the disagreement of semantic predictions inside one instance, which is brought in by the class-agnostic way of instance segmentation.
The strategy used in the proposed consensus-driven fusion is \textit{majority voting}.
For each predicted instance, the most appeared semantic label of its points determines
the semantic labels for all the points inside the instance.
This simple fusion strategy is not only efficient but could also revise and unify semantic predictions using instance information.

%% file: chaps/pics/04_02_ablation.tex
\begin{figure*}[ht]
    \begin{center}
        \includegraphics[width=1.0\linewidth]{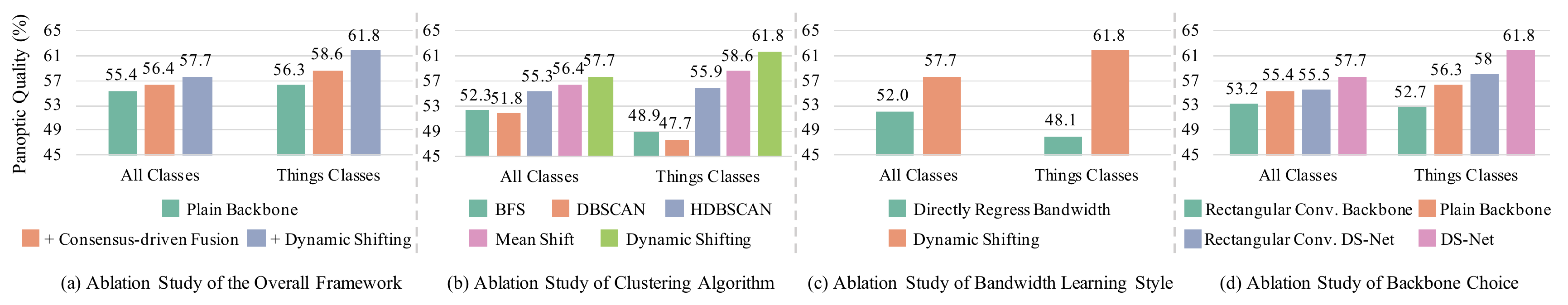}
    \end{center}
    \caption{\textbf{Ablation Study on the Validation Set of SemanticKITTI}. The proposed two modules both contributes to the final performance of the \nickname{}. The dynamic shifting module has advantages in clustering LiDAR point clouds. Weighting on bandwidth candidates is better than directly regressing bandwidth.}
    \label{fig:04_02_ablation}
\end{figure*}

%% file: chaps/03_04_4D.tex
\subsection{4D Panoptic LiDAR Segmentation} \label{3.4}

Based on the above proposed single version of the LiDAR-based panoptic segmentation method \nickname{}, we further extend it to the task of 4D panoptic LiDAR segmentation.

\noindent \textbf{To Extend from 3D to 4D.}
To extend from single frame panoptic segmentation to its 4D counterpart, the things IDs need to be consistent across frames.
In other words, for the same instance observed in multiple frames, the target is to assign the same IDs for them.
The trivial way is to append a tracking module to the instance segmentation branch to associate the predicted instance segments from previous and current frames.
However, such na\"ive stacking of modules will inevitably lead to compromised performance of tracking due to its dependence on the segmentation quality.
Moreover, the tracking module is hard to fully utilize the information provided by the consecutive LiDAR scans since it only processes the cropped partial observations.
It is challenging for the tracking module to extract distinctive features from incomplete sparse point clouds.
To fully utilize the temporal information from consecutive LiDAR scans, we propose to perform instance clustering in a temporally unified way, which is illustrated below.

\noindent \textbf{Temporally Unified Instance Clustering.} To ensure the consistency of the things IDs, we propose to use the temporally unified instance clustering to replace the explicit association.
The target of such clustering strategy is to jointly cluster all the points of the same instance from several frames to a single cluster.
Then we could naturally separate these points to different frames and allocate them the same instance ID.
To fit such clustering strategy into the bottom-up pipeline, we need to modify the targets of the center regression step and the following clustering module.
Assuming the point-level features of several consequent LiDAR frames are obtained, in the single frame version of the pipeline, the point-level features are used to regress the center of the instances.
However, in the multi-frame scenario, the traffic participants like cars and pedestrians would move to different positions in different frames.
Therefore, if we still follow the center regression target of the single frame version, there is great possibility of the same instance being clustered into multiple clusters because the regressed centers are too far apart due to high moving speed.
To avoid such problem, for the same instance, we propose to regress to the center of the overlapped point clouds from multiple frames, which can be formulated as
\begin{equation}
    C_{gt}(p_{id}) = \text{center}(\{p|p\in gt_{t}(id) \ldots gt_{t+i}(id)\}),
\end{equation}
where $p_{id}$ is the point that has the things ID of $id$, $gt_{t+i}(id)$ represents the set of points that have the things ID of $id$ and in frame $t+i$.
After the adjusted center regression, the following clustering step is performed on the overlapped regressed centers.
Unlike the 4D Volume Clustering proposed by \cite{aygun20214d}, our clustering process does not take the frame timestamp of each point into consideration, which means our method is frame agnostic.
Ideally, all the points of the same instance from several frames are clustered into a single cluster, which matches the target of the proposed temporally unified instance clustering.
To integrate the clustering strategy into the single frame version \nickname{}, we need to obtain the point-level features of several consequent LiDAR frames from the backbone.
Features from different frames need to be aware of the information of each other.
Otherwise, it is unreasonable to regress the centers of the overlapped point clouds.
There are two possible ways to achieve that.
The first one is to merge consequent LiDAR scans in the data level.
The second one is to merge the feature map of each individual frame right after the backbone.
From the perspective of computational efficiency, the first one is more efficient because only one 3D feature map need to be processed by the backbone.
As for the performance, we find out that the first one is also a better strategy through extensive experiments.
Therefore, based on the above analysis, we propose the 4D extension version of \nickname{}, which is illustrated below.

\noindent \textbf{4D extension of \nickname{}.}
The 4D extension version of \nickname{} (namely 4D-DS-Net) for the 4D panoptic segmentation is shown in Fig. \ref{fig:03_04_4d}.
We first align consequent LiDAR point clouds and overlap them to get the temporally fused LiDAR point clouds.
The fused LiDAR point clouds from frame $t$ to $t+i$ is defined as
\begin{align}
    P_{t:t+i} &= \{p|p\in P_t' \ldots P_{t+i}'\} \text{, where} \\
    P_{t+i}' &= ((P_{t+i}R_{t+i}^{-1} + T_{t+i})-T_t)R_t,
\end{align}
$R_{t+i}$ and $T_{t+i}$ represent the rotation matrix and translation vector of frame $t+i$.
The semantic segmentation branch predicts semantic labels for each point as that of the single version.
The instance segmentation branch produces temporally consistent IDs for each point, which is achieved by the temporally unified instance clustering proposed above.
Specifically, the foreground points are first regressed to the centers of the overlapped instances.
Then, the regressed centers are further clustered by the proposed dynamic shifting network in the frame agnostic way.
Such unified instance clustering step naturally associates the same instance across frames and saves the effort of tracking algorithms.
Once we ensure the consistency of things IDs in two consequent frames, it is trivial to achieve such consistency across the whole sequences, which gives the final 4D panoptic segmentation results.

%% file: chaps/tabs/04_02_tab_semkitti.tex
\begin{table*}[ht]
    \caption{LiDAR-based panoptic segmentation results on the validation set of SemanticKITTI. All results in [\%].}
    \begin{center}
    \small{
        \begin{tabular}{l|cccc|ccc|ccc|c}
            \Xhline{1pt}
            Method & \PQ & \PQda & \RQ & \SQ & \PQth & \RQth & \SQth & \PQst & \RQst & \SQst & \miou \\
            \hline\hline
            KPConv \cite{thomas2019kpconv} +
            PV-RCNN \cite{shi2020pv}              & 51.7         & 57.4          & 63.1          & \textbf{78.9} & 46.8          & 56.8          & \textbf{81.5} & \textbf{55.2} & \textbf{67.8} & \textbf{77.1} & 63.1          \\
            Cylinder3D \cite{zhou2020cylinder3d} +
            PV-RCNN \cite{shi2020pv} & 51.9 & 57.5 & 63.8 & 74.2 & 48.5 & 59.5 & 70.2 & 54.3 & 66.9 & \textbf{77.1} & 62.9 \\
            \hline
            PointGroup \cite{jiang2020pointgroup} & 46.1         & 54.0          & 56.6          & 74.6          & 47.7          & 55.9          & 73.8          & 45.0          & 57.1          & 75.1                      & 55.7          \\
            LPASD \cite{milioto2020iros}              & 36.5 & 46.1 & -    & -    & -    & 28.2 & -    & -    & -    & -    & 50.7 \\
            PanosterK \cite{gasperini2021panoster}    & 55.6 & -    & 66.8 & 79.9 & 56.6 & 65.8 & -    & -    & -    & -    & 61.1 \\
            Panoptic-PolarNet \cite{zhou2021panoptic} & \textbf{59.1} & \textbf{64.1} & \textbf{70.2} & \textbf{78.3} & \textbf{65.7} & \textbf{74.7} & \textbf{87.4} & 54.3 & 66.9 & 71.6 & \textbf{64.5} \\
            \hline
            \nickname{} & 57.7 & 63.4 & 68.0 & 77.6 & 61.8 & 68.8 & 78.2 & \textbf{54.8} & \textbf{67.3} & \textbf{77.1} & 63.5 \\
            \Xhline{1pt}
        \end{tabular}
    }
    \end{center}
    \label{tab:semkitti_val}
\end{table*}

\begin{table*}[ht]
    \caption{LiDAR-based panoptic segmentation results on the test set of SemanticKITTI. All results in [\%].}
    \begin{center}
    \small{
        \begin{tabular}{l|cccc|ccc|ccc|c}
            \Xhline{1pt}
            Method & \PQ & \PQda & \RQ & \SQ & \PQth & \RQth & \SQth & \PQst & \RQst & \SQst & \miou \\
            \hline\hline
            KPConv \cite{thomas2019kpconv} +
            PointPillars \cite{lang2019pointpillars} & 44.5          & 52.5          & 54.4          & 80.0          & 32.7          & 38.7          & 81.5          & 53.1          & 65.9          & 79.0                  & 58.8          \\
            RangeNet++ \cite{milioto2019rangenet++} +
            PointPillars \cite{lang2019pointpillars} & 37.1          & 45.9          & 47.0          & 75.9          & 20.2          & 25.2          & 75.2          & 49.3          & 62.8          & 76.5                  & 52.4          \\
            KPConv \cite{thomas2019kpconv} +
            PV-RCNN \cite{shi2020pv}                 & 50.2          & 57.5          & 61.4          & 80.0          & 43.2          & 51.4          & 80.2          & 55.9 & 68.7 & \textbf{79.9} & \textbf{62.8} \\
            \hline
            LPASD \cite{milioto2020iros}              & 38.0          & 47.0          & 48.2          & 76.5          & 25.6          & 31.8          & 76.8          & 47.1          & 60.1          & 76.2                  & 50.9 \\
            PanosterK \cite{gasperini2021panoster}    & 52.7 & 59.9 & 64.1 & 80.7 & 49.4 & 58.5 & 83.3 & 55.1 & 68.2 & 78.8 & 59.9 \\
            4D-PLS \cite{aygun20214d}                 & 50.3 & 57.8 & 61.0 & 81.6 & -    & -    & -    & -    & -    & -    & 61.3 \\
            Panoptic-PolarNet \cite{zhou2021panoptic} & 54.1 & 60.7 & 65.0 & 81.4 & 53.3 & 60.6 & 87.2 & 54.8 & 68.1 & 77.2 & 59.5 \\
            \hline
            \nickname{} & \textbf{55.9} & \textbf{62.5} & \textbf{66.7} & \textbf{82.3} & \textbf{55.1} & \textbf{62.8} & \textbf{87.2} & \textbf{56.5} & \textbf{69.5} & 78.7 & 61.6 \\
            \Xhline{1pt}
        \end{tabular}
    }
    \end{center}
    \label{tab:semkitti_test}
\end{table*}

%% file: chaps/04_01_setting.tex
We conduct experiments on two large-scale datasets: SemanticKITTI \cite{behley2019semantickitti} and
nuScenes \cite{nuscenes2019}. In addition, we evaluate our extension of 4D panoptic segmentation on SemanticKITTI.

\noindent\textbf{SemanticKITTI.}
SemanticKITTI is the first dataset that presents the challenge of LiDAR-based panoptic segmentation
and provides the benchmark\cite{behley2020benchmark}.
SemanticKITTI contains 23,201 frames for training and 20,351 frames for testing.
There are 28 annotated semantic classes which are remapped to 19 classes for the LiDAR-based panoptic segmentation task,
among which 8 classes are \things{} classes, and 11 classes are \stuff{} classes.
Each point is labeled with a semantic label and an temporally consistent instance ID which will be set to 0 if the point belongs to \stuff{} classes.

\noindent\textbf{nuScenes.}
In order to demonstrate the generalizability of \nickname{}, we construct another LiDAR-based panoptic
segmentation dataset from nuScenes.
With the point-level semantic labels from the newly released nuScenes \textit{lidarseg} challenge and the bounding
boxes provided by the detection task, we could generate instance labels by assigning instance IDs to
points inside bounding boxes.
Following the definition of the nuScenes \textit{lidarseg} challenge, we mark 10 foreground classes as \things{} classes
and 6 background classes as \stuff{} classes out of all 16 semantic classes.
The training and validation set has 28,130 and 6,019 frames.

\noindent\textbf{Evaluation Metrics of LiDAR-based Panoptic Segmentation.}
As defined in \cite{behley2020benchmark}, the evaluation metrics of LiDAR-based panoptic segmentation are the
same as that of image panoptic segmentation defined in \cite{kirillov2019panoptic} including Panoptic Quality
(PQ), Segmentation Quality (SQ) and Recognition Quality (RQ) which are calculated across all classes. For each class, the PQ, SQ and RQ are defined as
\begin{equation}
    \text{PQ} = \underbrace{\frac{\sum_{(i, j)\in TP}\text{IoU}(i, j)}{|TP|}}_\text{SQ} \times \underbrace{\frac{|TP|}{|TP| + \frac{1}{2}|FP| + \frac{1}{2}|FN|}}_\text{RQ}.
\end{equation}
The above three metrics are also calculated separately on \things{} and \stuff{} classes which give
\PQth{}, \SQth{}, \RQth{}, and \PQst{}, \SQst{}, \RQst{}.
\PQda{} is defined by swapping \PQ{} of each \stuff{} class to its IoU then averaging over all classes.
In addition, mean IoU (mIoU) is also used to evaluate the quality of the sub-task of
semantic segmentation.

\noindent\textbf{Evaluation Metrics of 4D Panoptic LiDAR Segmentation.}
Several metrics are proposed by previous video panoptic segmentation works \cite{kim2020video, qiao2021vip, aygun20214d}. Among them, we choose to use LSTQ (LIDAR Segmentation and Tracking Quality) \cite{aygun20214d} as the evaluation metrics for 4D Panoptic Segmentation, which is defined as
\begin{equation}
    \text{LSTQ} = \sqrt{\underbrace{\frac{1}{C}\sum_{c=1}^{C} \text{IoU}(c)}_\text{\scls} \times \underbrace{\frac{1}{T}\sum_{t\in T}\frac{\sum_{s\in S}\text{TPA}(s, t)\text{IoU}(s, t)}{|gt_{id}(t)|} }_\text{\sass} },
\end{equation}
where \scls{} and \sass{} reflects the segmentation and tracking quality respectively. TPA (True Positive Association) is defined as $TPA(i, j) = |pr(i)\cap gt(j)|$, which represents the number of intersection between points that are predicted as $i$ and the ground truth points that have the id of $j$.

\noindent\textbf{Implementation Details of Backbone.}
For both datasets, each input point is represented as a $4$ dimension vector including XYZ coordinates and the intensity.
The backbone voxelizes a single frame to $480\times 360\times 32$ voxels under the cylindrical coordinate
system.
For that we should not use the information of bounding boxes in this segmentation task, the ground truth
center of each instance is approximated by the center of its tight box that parallel to axes which makes a better
approximation than the mass centers of the incomplete point clouds.
The bandwidth of the Mean Shift used in our backbone method is set to $1.2$.
Adam solver is utilized to optimize the network.
The minimum number of points in a valid instance is set to 50 for SemanticKITTI and 5 for nuScenes.

\noindent\textbf{Implementation Details of Dynamic Shifting.}
The number of the FPS downsampled points in the dynamic shifting module is set to $10000$.
The final heuristic clustering algorithm used in the dynamic shifting module is Mean Shift with $0.65$ bandwidth for
SemanticKITTI and BFS with $1.2$ radius for nuScenes.
Bandwidth candidates are set to $0.2$, $1.7$ and $3.2$ for both datasets.
The number of Iterations is set to $4$ for both datasets.
We train the network with the learning rate of $0.002$, epoch of $50$ and batch size of $4$ on four Geforce GTX 1080Ti.
The dynamic shifting module only takes 3-5 hours to train on top of a pretrained backbone.

\noindent\textbf{Implementation Details of 4D-DS-Net.}
Two consequent LiDAR scans are aligned and overlapped for the training and inference of 4D panoptic segmentation. The number of the FPS downsampled points in the dynamic shifting module is set to $20000$. Other hyper-parameters are the same as its single version counterpart.

%% file: chaps/tabs/04_03_tab_nuscenes.tex
\begin{table*}[ht]
    \caption{LiDAR-based panoptic segmentation results on the validation set of nuScenes. All results in [\%].}
    \begin{center}
    \small{
        \begin{tabular}{l|cccc|ccc|ccc|c}
            \Xhline{1pt}
            Method       & \PQ  & \PQda & \RQ  & \SQ  & \PQth & \RQth & \SQth & \PQst & \RQst & \SQst & \miou \\
            \hline\hline
            Cylinder3D \cite{zhou2020cylinder3d} +
            PointPillars \cite{lang2019pointpillars} & 36.0          & 44.5          & 43.0          & 83.3          & 23.3          & 27.0          & 83.7          & 57.2          & 69.6          & 82.7                    & 52.3          \\
            Cylinder3D \cite{zhou2020cylinder3d} +
            SECOND \cite{yan2018second}              & 40.1          & 48.4          & 47.3          & \textbf{84.2} & 29.0          & 33.6          & \textbf{84.4} & 58.5          & 70.1          & 83.7                    & 58.5          \\
            \hline
            \nickname{}                              & \textbf{42.5} & \textbf{51.0} & \textbf{50.3} & 83.6          & \textbf{32.5} & \textbf{38.3} & 83.1          & \textbf{59.2} & \textbf{70.3} & \textbf{84.4} & \textbf{70.7} \\
            \Xhline{1pt}
        \end{tabular}
    }
    \end{center}
    \label{tab:nus_val}
\end{table*}

%% file: chaps/04_02_semkitti.tex
\subsection{Ablation Study}\label{ablation}

\noindent\textbf{Ablation on Overall Framework.}
To study on the effectiveness of the proposed modules, we sequentially add consensus-driven fusion
module and dynamic shifting module to the bare backbone.
The corresponding \PQ{} and \PQth{} are reported in Fig. \ref{fig:04_02_ablation} (a) which shows
that both modules contribute to the performance of \nickname{}.
The novel dynamic shifting module mainly boosts the performance of instance segmentation which are indicated by
\PQth{} where the \nickname{} outperforms the backbone (with fusion module) by 3.2\% in validation split.

\noindent\textbf{Ablation on Clustering Algorithms.}
In order to validate our previous analyses of clustering algorithms, we swap the dynamic shifting module for four
other widely-used heuristic clustering algorithms: BFS, DBSCAN, HDBSCAN, and Mean Shift.
The results are shown in Fig. \ref{fig:04_02_ablation} (b).
Consistent with our analyses in Sec. \ref{3.2}, the density-based clustering algorithms (\eg BFS, DBSCAN, HDBSCAN)
perform badly in terms of \PQ{} and \PQth{} while Mean Shift leads to the best results among the heuristic algorithms.
Moreover, our dynamic shifting module shows the superiority over all four heuristic clustering algorithms.

\noindent\textbf{Ablation on Bandwidth Learning Styles.}
In the dynamic shifting module, it is natural to directly regress bandwidth for each point as mentioned in Sec. \ref{3.2}.
However, as shown in the Fig. \ref{fig:04_02_ablation} (c), direct regression is hard to optimize in this
case because the learning target is not straightforward.
It is difficult to determine the best bandwidth for each point, and therefore impractical to directly apply supervision
on the regressed bandwidth.
Therefore, it is easier for the network to choose from and combine several bandwidth candidates.

\noindent\textbf{Ablation on Backbone Choice.}
To demonstrate that the dynamic shifting module can apply to different backbones, we report the performance of a rectangular convolution version of plain backbone and DS-Net Fig. \ref{fig:04_02_ablation} (d).
An improvement of 2.3\% in terms of \PQ{} on both the rectangular convolution version and cylinder convolution version of the plain backbone is achieved on the validation set of SemanticKITTI, which shows the generalizability of the proposed dynamic shifting module.

\subsection{Evaluation Comparisons on SemanticKITTI}

\noindent\textbf{Comparison Methods.}
Since its one of the first attempts on LiDAR-based panoptic segmentation,
we provide several strong baseline results in order to validate the effectiveness of \nickname{}.
As proposed in \cite{behley2020benchmark}, one good way of constructing strong baselines is to take the results from
semantic segmentation methods and detection methods, and generate panoptic segmentation results by assigning instance
IDs to all points inside predicted bounding boxes.
\cite{behley2020benchmark} has provided the combinations of KPConv \cite{thomas2019kpconv} +
PointPillars \cite{lang2019pointpillars}, and RangeNet++ \cite{milioto2019rangenet++} +
PointPillars \cite{lang2019pointpillars}.
To make the baseline stronger, we combine KPConv \cite{thomas2019kpconv} with
PV-RCNN \cite{shi2020pv} which is the state-of-the-art 3D detection method.
In addition to the above baselines, we also adapt the state-of-the-art indoor
instance segmentation method PointGroup \cite{jiang2020pointgroup} using the official released codes to experiment on SemanticKITTI.
We also compare with recent LiDAR-based Panoptic Segmentation works: LPASD \cite{milioto2020iros}, PanosterK \cite{gasperini2021panoster} and Panoptic-PolarNet \cite{zhou2021panoptic}.

\noindent\textbf{Evaluation Results.}
Table \ref{tab:semkitti_val} and \ref{tab:semkitti_test} shows that the \nickname{}
outperforms most existing methods in both validation and test splits by a large margin.
It is worth noting that PointGroup \cite{jiang2020pointgroup} performs poorly on the LiDAR point clouds which shows that indoor solutions are not suitable for challenging LiDAR point clouds.
In test split, the \nickname{} outperforms state-of-the-art method Panoptic-PolarNet \cite{zhou2021panoptic} by 1.8\% in both \PQ{} and \PQth{}.

%% file: chaps/04_03_nuscenes.tex
\subsection{Evaluation Comparisons on nuScenes}

\noindent\textbf{Comparison Methods.}
Similarly, two strong semantic segmentation + detection baselines are provided for comparison on nuScenes.
The semantic segmentation method is Cylinder3D \cite{zhou2020cylinder3d} and the detection
methods are SECOND \cite{yan2018second} and PointPillars \cite{lang2019pointpillars}.
For fair comparison, the detection networks are trained using single frames on nuScenes.
The point-wise semantic predictions and predicted bounding boxes are merged in the following steps.
First all points inside each bounding box are assigned a unique instance IDs across the frame.
Then to unify the semantic predictions inside each instance, we assign the class labels of bounding
boxes predicted by the detection network to corresponding instances.

\noindent\textbf{Evaluation Results.}
As shown in Table \ref{tab:nus_val}, our \nickname{} outperforms the best baseline method in most metrics.
Especially, we surpass the best baseline method by 2.4\% in \PQ{} and 3.5\% in \PQth{}.
Unlike SemanticKITTI, nuScenes is featured as extremely sparse point clouds in single frames which
adds even more difficulties to panoptic segmentation.
The results validate the generalizability and the effectiveness of our \nickname{}.

%% file: chaps/04_04_4D.tex
\input{chaps/tabs/04_04_4D_val}
\input{chaps/tabs/04_04_4D_test}
\begin{table}[t]
    \caption{Results of Single Frame \PQ{} Evaluation of 4D-\nickname{} on the validation set of SemanticKITTI.}
    \begin{center}
    \small{
        \begin{tabular}{c|cccc|c}
            \Xhline{1pt}
            Name & \PQ & \PQda & \PQth & \PQst & \miou \\
            \hline\hline
            \nickname{} & 57.7 & 63.4 & 61.8 & 54.8 & 63.5 \\
            \nickname{} + Feat. Fus. & 58.6 & 63.9 & 63.4 & 55.0 & 63.7 \\
            4D-\nickname{}  & \textbf{59.5} & \textbf{64.5} & \textbf{64.4} & \textbf{55.9} & \textbf{64.8} \\
            \Xhline{1pt}
        \end{tabular}
    }
    \end{center}
    \label{tab:multif}
\end{table}

\subsection{4D Panoptic LiDAR Segmentation Results}

\noindent \textbf{Comparison Methods.} Since the task is fairly new, we choose to compare with the first work \cite{aygun20214d} that proposes this task, denoted as 4D-PLS, and several `Semantic Segmentation + 3D Object Detection + Tracking' assembled baseline methods. Besides, we also construct a baseline method `\nickname{} + Tracking' by appending a tracking module \cite{kim2020video} to the instance segmentation branch. As discussed in Sec. \ref{3.4}, we also implement the feature map fusion on top of \nickname{}, namely `\nickname{} + Feat. Fus.'. Specifically, we perform a max pooling operation on the aligned 3D feature maps extracted from consecutive LiDAR frames by the backbone. Then the fused feature map is fed to semantic and instance branches. We refer to our proposed method as `4D-DS-Net'.

\noindent \textbf{Evaluation Results.} As shown in Table \ref{tab:4d_val} and \ref{tab:4d_test}, our proposed method surpasses all baseline methods and the state-of-the-art method 4D-PLS \cite{aygun20214d} in terms of the main metric \lstq{} in both validation and test sets. Moreover, improvements of 9.4\% and 1.5\% in respect of \sass{} and \scls{} on the test set are achieved by \nickname{}, which shows that both the tracking and segmentation performance contribute to the overall 5.4\% improvement of \lstq{}. The proposed `4D-\nickname{}' also surpasses `\nickname{} + Tracking' by 2.1\% in terms of \lstq{} on the validations set, which proves that simply stacking modules is hard to fully utilize the temporal information as mentioned in Sec. \ref{3.4}. Moreover, `4D-DS-Net' surpasses `\nickname{} + Feat. Fus.' by a small margin of 0.2\% in terms of \lstq{}. However, the amount of memory and computation overhead of `\nickname{} + Feat. Fus.' compared to that of `4D-DS-Net' still justifies our preference of data-level fusion.

\noindent\textbf{4D Panoptic Segmentation Improves the Single Frame \PQ{} Evaluation.}
We also evaluate the single frame metrics using the 4D version of \nickname{} on the validation set of SemanticKITTI. As shown in Table \ref{tab:multif}, the 4D version of \nickname{} surpasses the single frame \nickname{} by 1.8\% in terms of \PQ{}. It shows that the temporal information can largely enrich the semantic information extracted by the backbone and therefore improves the overall performance. The improved single frame segmentation quality also explains the better performance in the task of 4D panoptic LiDAR segmentation. Moreover, `4D-DS-Net' also outperforms `\nickname{} + Feat. Fus.' by 0.9\% in terms of \PQ{}, which shows the superiority of data-level fusion over simple feature-level fusion. Of course, more complex feature fusion could be designed and has the potential of outperforming data-level fusion. But with minimal memory and computation overhead, data-level fusion is the first choice here.

%% file: chaps/tabs/04_04_4D_val.tex
\begin{table}[t]
    \caption{4D panoptic LiDAR segmentation results on the validation set of SemanticKITTI. All results in [\%]. RN: RangeNet++ \cite{milioto2019rangenet++}; KP: KPConv \cite{thomas2019kpconv}; PP: PointPillars \cite{lang2019pointpillars}; MOT: Multi-Object Tracking \cite{weng20203d}; SFP: Scene Flow based Propagation \cite{mittal2020just}.}
    \begin{center}
    \small{
        \begin{tabular}{l|ccc|cc}
            \Xhline{1pt}
            Method & \lstq & \sass & \scls & \ioust & \iouth \\
            \hline\hline
            RN + PP + MOT & 43.8 & 36.3 & 52.8 & 60.5 & 42.2 \\
            KP + PP + MOT & 46.3 & 37.6 & 57.0 & 64.2 & 54.1 \\
            RN + PP + SFP & 43.4 & 35.7 & 52.8 & 60.5 & 42.2 \\
            KP + PP + SFP & 46.0 & 37.1 & 57.0 & 64.2 & 54.1 \\
            \hline
            4D-PLS \cite{aygun20214d} & 62.8 & 65.1 & 60.5 & \textbf{65.4} & 61.3 \\
            \hline
            \nickname{} + Tracking & 65.9 & 68.4 & 63.1 & 64.0 & 61.9 \\
            \nickname{} + Feat. Fus. & 67.8 & \textbf{72.1} & 63.7 & 64.2 & 63.1 \\
            4D-\nickname{} & \textbf{68.0} & 71.3 & \textbf{64.8} & 64.5 & \textbf{65.30} \\
            \Xhline{1pt}
        \end{tabular}
    }
    \end{center}
    \label{tab:4d_val}
\end{table}

%% file: chaps/tabs/04_04_4D_test.tex
\begin{table}[t]
    \caption{4D panoptic LiDAR segmentation results on the test set of SemanticKITTI. All results in [\%].
    RN: RangeNet++ \cite{milioto2019rangenet++}; KP: KPConv \cite{thomas2019kpconv}; PP: PointPillars \cite{lang2019pointpillars}; MOT: Multi-Object Tracking \cite{weng20203d}; SFP: Scene Flow based Propagation \cite{mittal2020just}.
    }
    \begin{center}
    \small{
        \begin{tabular}{l|ccc|cc}
            \Xhline{1pt}
            Method & \lstq & \sass & \scls & \ioust & \iouth \\
            \hline\hline
            RN + PP + MOT & 35.5 & 24.1 & 52.4 & 64.5 & 35.8 \\
            KP + PP + MOT & 38.0 & 25.9 & 55.9 & 66.9 & 47.7 \\
            RN + PP + SFP & 34.9 & 23.3 & 52.4 & 64.5 & 35.8 \\
            KP + PP + SFP & 38.5 & 26.6 & 55.9 & 66.9 & 47.7 \\
            \hline
            4D-PLS \cite{aygun20214d} & 56.9 & 56.4 & 57.4 & \textbf{66.9} & \textbf{51.6} \\
            \hline
            4D-\nickname{} & \textbf{62.3} & \textbf{65.8} & \textbf{58.9} & 65.6 & 49.8 \\
            \Xhline{1pt}
        \end{tabular}
    }
    \end{center}
    \label{tab:4d_test}
\end{table}

%% file: chaps/04_05_analyze.tex
\subsection{Further Analysis}
\noindent\textbf{Robust to Parameter Settings.}
As shown in Table \ref{tab:ab_bc}, six sets of bandwidth candidates are set for independent
training and the corresponding results are reported.
The stable results show that \nickname{} is robust to different parameter settings as long
as the picked bandwidth candidates are comparable to the instance sizes.
Unlike previous heuristic clustering algorithms that require massive parameter adjustment,
\nickname{} can automatically adjust to different instance sizes and point distributions and
remains stable clustering quality.
\input{chaps/tabs/04_02_tab_ab_bc}

\noindent\textbf{Interpretable Learned Bandwidths.}
By averaging the bandwidth candidates weighted by the learned weights, the learned bandwidths for every
points could be approximated.
The average learned bandwidths of different classes are shown in Fig.
\ref{fig:04_04_class_bandwidth}.
The average learned bandwidths are roughly proportional to the instance sizes of corresponding classes, which is
consistent with the expectation that dynamic shifting can dynamically adjust to different instance sizes.
\input{chaps/pics/04_04_class_bandwidth}
\input{chaps/pics/04_04_diffpart_bandwidth}

\noindent\textbf{Visualization of Dynamic Shifting Iterations.}
As visualized in Fig. \ref{fig:04_04_diffpart_bandwidth}, the black points are
the original point clouds of different instances including person, bicyclist and car.
The seeding points are colored in spectral colors where the redder points represents higher learned bandwidth
and bluer points represents lower learned bandwidth.
The seeding points farther away from the instance centers tend to learn higher
bandwidths in order to quickly converge.
While the well-learned regressed points tend to have lower bandwidths to maintain their positions.
After four iterations, the seeding points have converged around the instance centers.

\noindent\textbf{Learned Bandwidths of Different Iterations.}
The average learned bandwidths of different iterations are shown in Fig. \ref{fig:04_04_iteration_bandwidth}.
As expected, as the iteration rounds grow, points of the same instance gather tighter which usually require
smaller bandwidths.
After four iterations, learned bandwidths of most classes have dropped to 0.2, which is the lowest they
can get, meaning that four iterations are enough for \things{} points to converge to cluster centers, which
further validates the conclusion made in the last paragraph.
\input{chaps/pics/04_04_iteration_bandwidth}

%% file: chaps/tabs/04_02_tab_ab_bc.tex
\begin{table}[h]
    \caption{Results of different bandwidth candidates settings. All results in [\%].}
    \begin{center}
    \small{
        \begin{tabular}{l|cccc|c}
            \Xhline{1pt}
            \makecell{Bandwidth\\ Candidates (m)} & \PQ & \PQda & \RQ & \SQ & \miou \\
            \hline\hline
            0.2, 1.1, 2.0 & 57.4          & 63.0          & 67.7          & 77.4          & \textbf{63.7} \\
            0.2, 1.3, 2.4 & 57.5          & 63.1          & 67.7          & \textbf{77.6} & 63.5          \\
            0.2, 1.5, 2.8 & 57.6          & 63.2          & 67.8          & \textbf{77.6} & \textbf{63.7} \\
            0.2, 1.7, 3.2 & \textbf{57.7} & \textbf{63.4} & \textbf{68.0} & \textbf{77.6} & 63.5          \\
            0.2, 1.9, 3.6 & \textbf{57.7} & 63.3          & 67.9          & \textbf{77.6} & 63.4          \\
            0.2, 2.1, 4.0 & 57.4          & 63.1          & 67.7          & 77.5          & 63.3          \\
            \Xhline{1pt}
        \end{tabular}
    }
    \end{center}
    \label{tab:ab_bc}
\end{table}

%% file: chaps/pics/04_04_class_bandwidth.tex
\begin{figure}[ht]
    \begin{center}
        \includegraphics[width=1.0\linewidth]{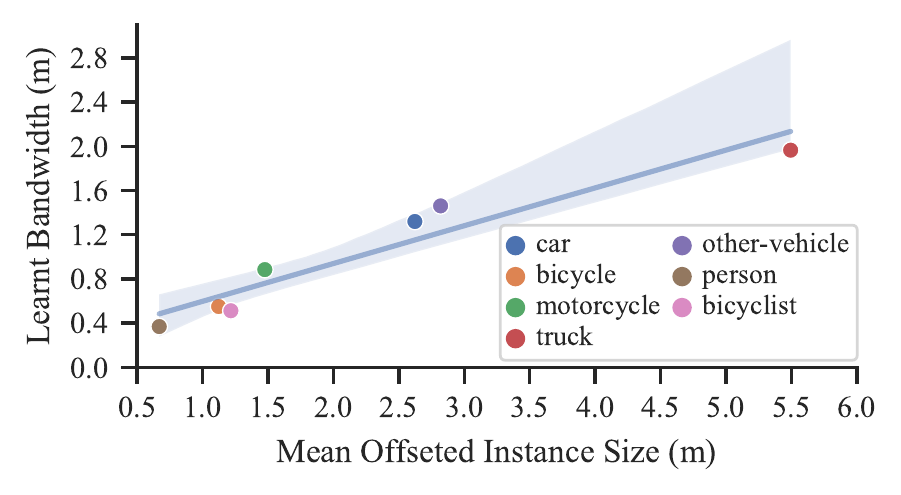}
    \end{center}
    \caption{\textbf{Proportional Relationship Between Sizes and the Learned Bandwidths.} The $x$-axis represents the class-wise average size of regressed centers of instances while the $y$-axis stands for the average learned bandwidth of different \things{} classes.}
    \label{fig:04_04_class_bandwidth}
\end{figure}

%% file: chaps/pics/04_04_diffpart_bandwidth.tex
\begin{figure}[ht]
    \begin{center}
        \includegraphics[width=1.0\linewidth]{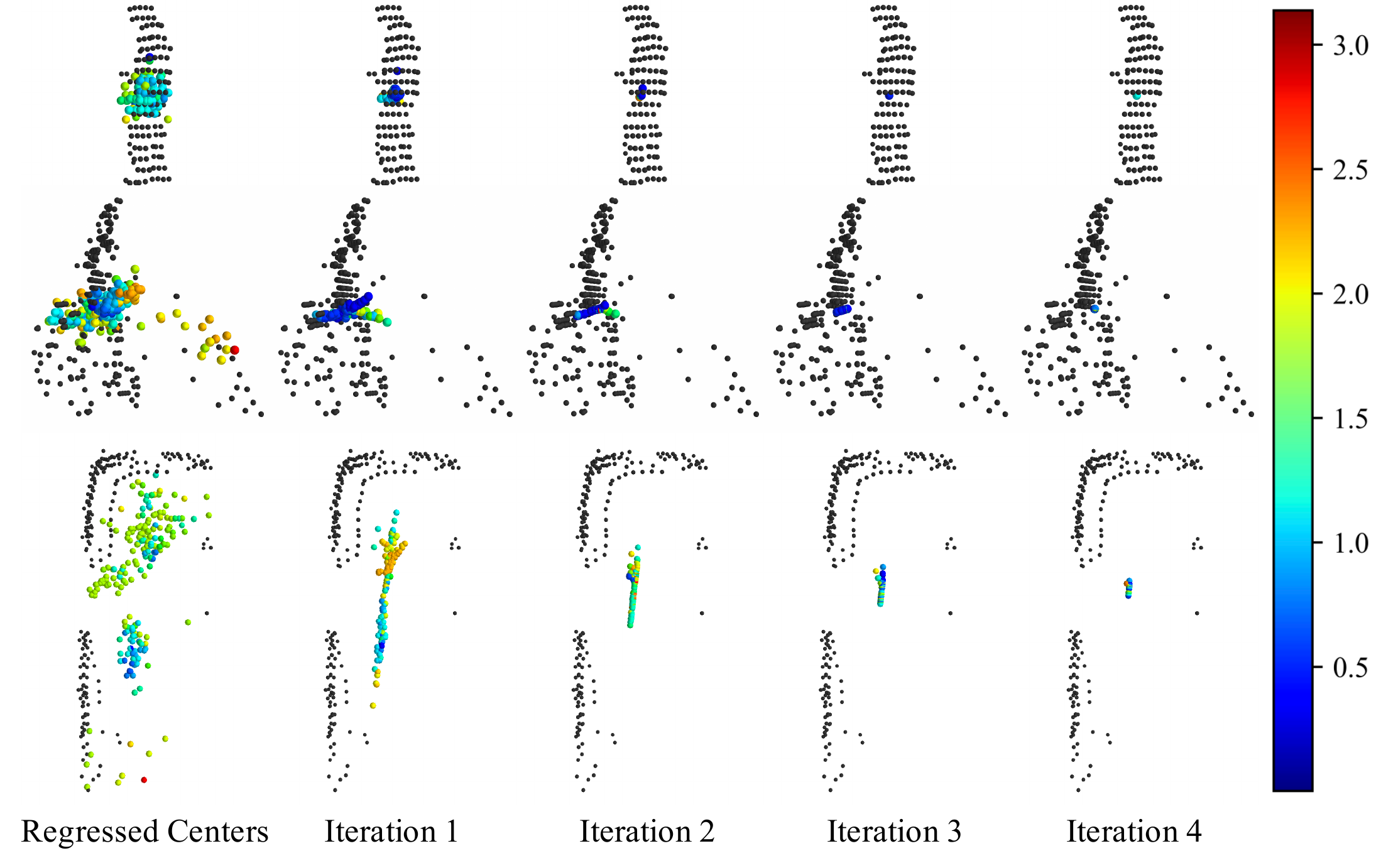}
    \end{center}
    \caption{\textbf{Visualization of Dynamic Shifting Iterations.} The black points are the original LiDAR point clouds of instances. The colored points are seeding points. From left to right, with the iteration number increases, the seeding points converge to cluster centers.}
    \label{fig:04_04_diffpart_bandwidth}
\end{figure}

%% file: chaps/pics/04_04_iteration_bandwidth.tex
\begin{figure}[ht]
    \begin{center}
        \includegraphics[width=1.0\linewidth]{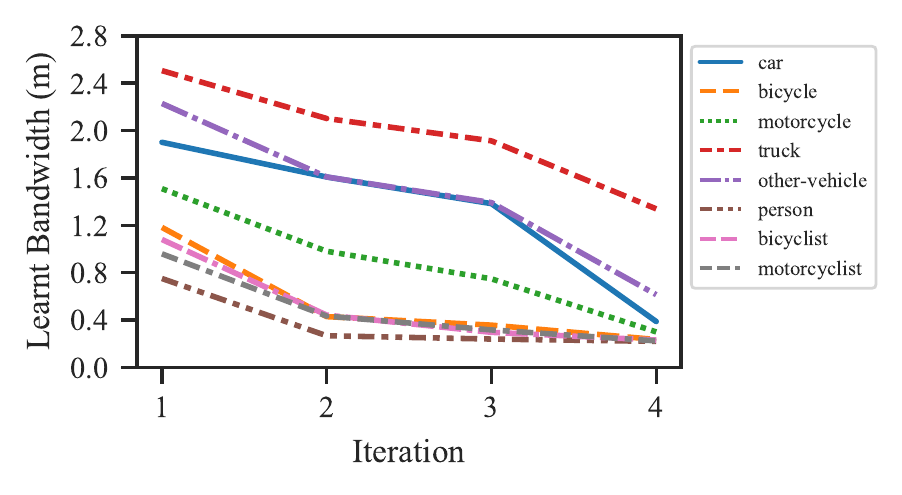}
    \end{center}
    \caption{\textbf{Relationship Between Iterations and the Learned Bandwidths.} With number of iteration increases, the learned bandwidth decreases. At the 4th iteration, the learned bandwidths of most classes drop near the lower limit.}
    \label{fig:04_04_iteration_bandwidth}
\end{figure}

%% file: chaps/05_conc.tex
With the goal of providing holistic perception for autonomous driving, we are one of the first to address the
task of LiDAR-based panoptic segmentation.
In order to tackle the challenge brought by the non-uniform distributions of LiDAR point clouds, we propose the
novel \nickname{} which is specifically designed for effective panoptic segmentation of LiDAR point clouds.
Our \nickname{} adopts strong baseline design which provides strong support for the consensus-driven fusion module
and the novel dynamic shifting module.
The novel dynamic shifting module adaptively shifts regressed centers of instances with different density and varying sizes.
The consensus-driven fusion efficiently unifies semantic and instance results into panoptic segmentation results.
The \nickname{} outperforms all strong baselines on both SemanticKITTI and nuScenes.
Moreover, we extend the single version of \nickname{} to the new task of LiDAR-based 4D panoptic segmentation and demonstrate state-of-the-art performance on SemanticKITTI.
Further analyses show the robustness of the dynamic shifting module and the interpretability of the learned bandwidths.

%% file: utils/biography.tex
\begin{IEEEbiography}[{\includegraphics[width=1in,height=1.25in,clip,keepaspectratio]{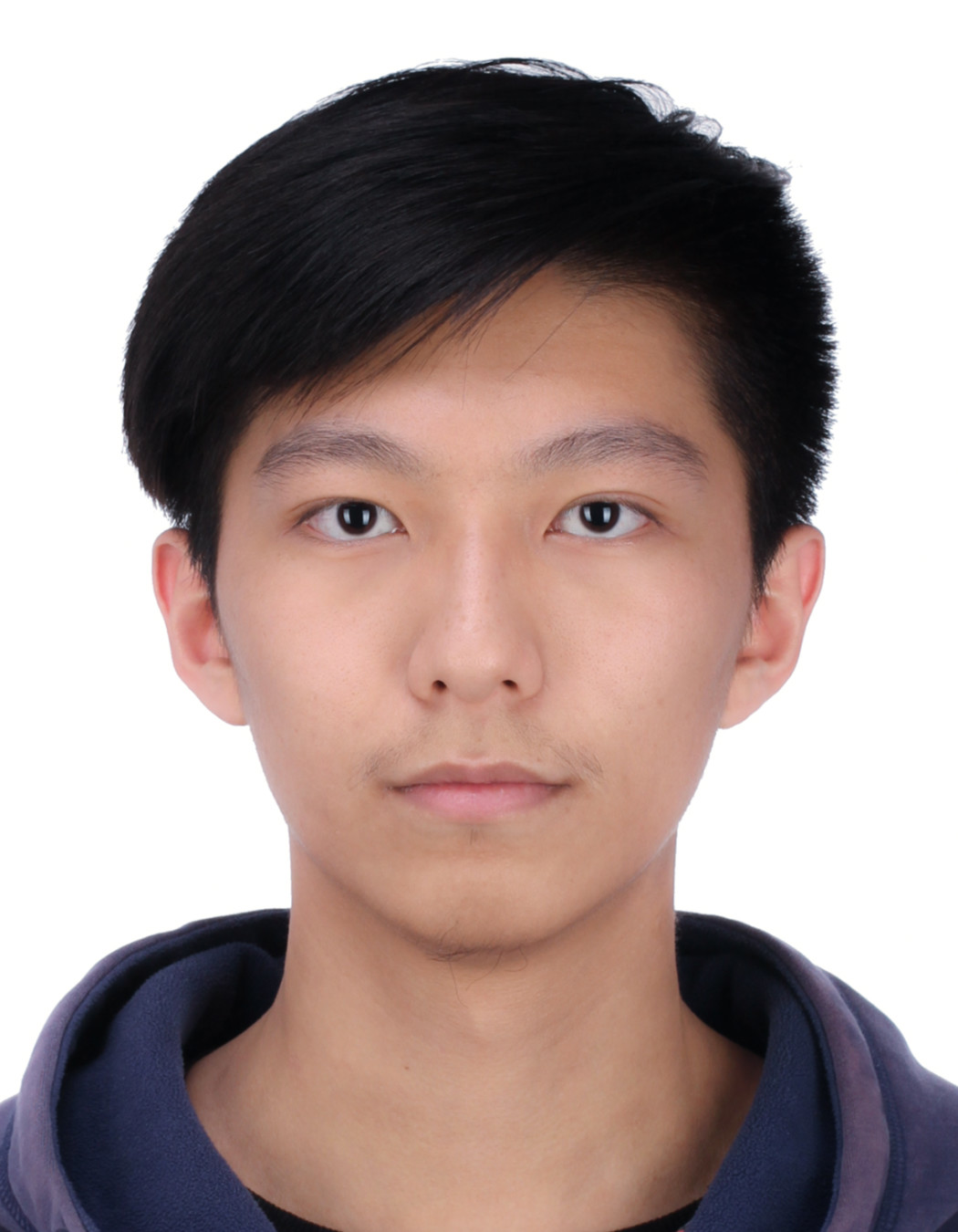}}]{Fangzhou Hong} received the BEng degree in Software Engineering from Tsinghua University, China, in 2020. He is currently a Ph.D. student in the School of Computer Science and Engineering at Nanyang Technological University. His research interests lie on the computer vision and deep learning. Particularly he is interested in 3D representation learning.
\end{IEEEbiography}
\begin{IEEEbiography}[{\includegraphics[width=1in,height=1.25in,clip,keepaspectratio]{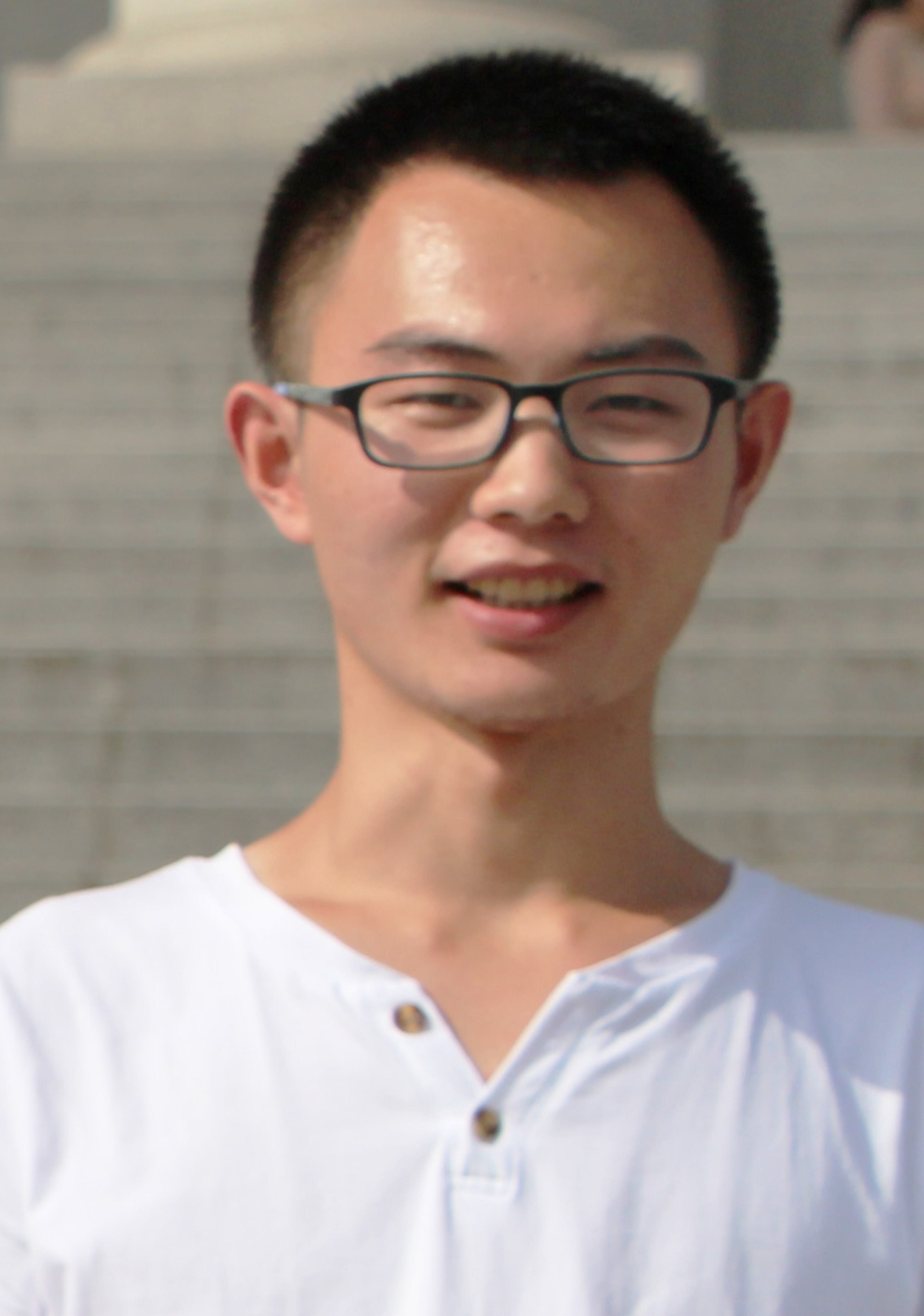}}]{Hui Zhou} received the bachelors and masters degree at university of science and electronic technology of china(UESTC) in 2015, 2018. He is currently a research scientist for autonomous driving in Sensetime Research. His research interests include computer vision and machine learning. 
\end{IEEEbiography}
\begin{IEEEbiography}[{\includegraphics[width=1in,height=1.25in,clip,keepaspectratio]{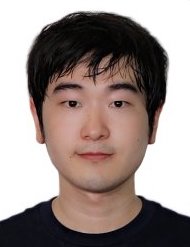}}]{Xinge Zhu} received the BEng degree in computer science from Shandong University, China, in 2015. He is working toward the PhD degree in The Chinese University of Hong Kong, under the supervision of Professor Dahua Lin. His research interests lie on the computer vision and machine learning. Particularly he is interested in 3D perception for autonomous vehicles, including 3D detection and 3D segmentation. 
\end{IEEEbiography}
\begin{IEEEbiography}[{\includegraphics[width=1in,height=1.25in,clip,keepaspectratio]{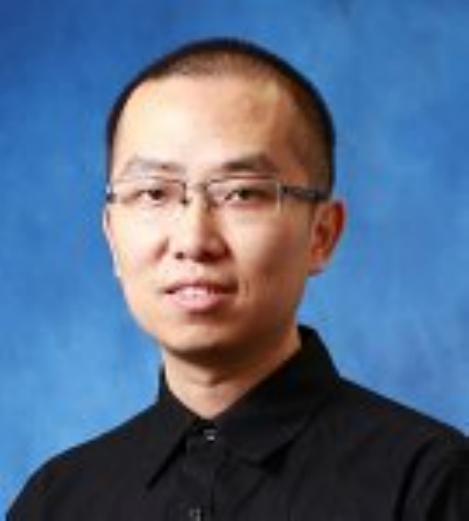}}]{Hongsheng Li} is an assistant professor in the Department of Electronic Engineering at the Chinese University of Hong Kong. In 2013-2015, he was an associate professor in the School of Electronic Engineering at University of Electronic Science and Technology of China. He received the bachelor’s degree in Automation from East China University of Science and Technology in 2006, and the doctorate degree in Computer Science from Lehigh University, United States in 2012.  He has published over 70 papers in premier conferences on computer vision and machine learning, including CVPR, ICCV, ECCV, NeurIPS, ICLR, and AAAI. He received the 2020 IEEE CAS Society Outstanding Young Author Award. He won the first place in Object Detection from Videos (VID) track of ImageNet challenge 2016 as the team leader and 2015 as a team co-leader. He is an associate editor of Neurocomputing and serves as an area chair of NeurIPS 2021. His research interests include computer vision, machine learning, and medical image analysis.
\end{IEEEbiography}
\begin{IEEEbiography}[{\includegraphics[width=1in,height=1.25in,clip,keepaspectratio]{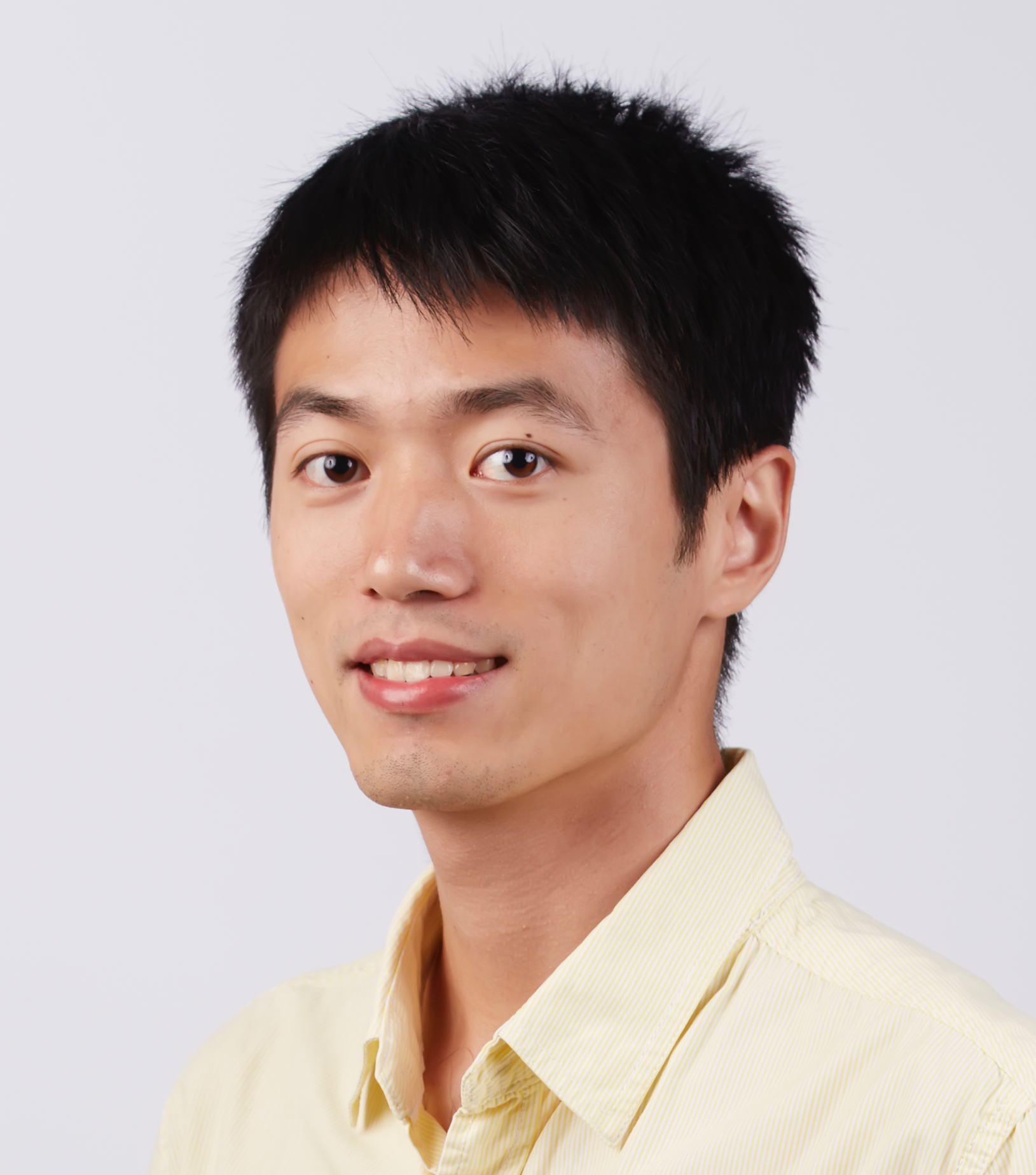}}]{Ziwei Liu} is a Nanyang Assistant Professor at School of Computer Science and Engineering (SCSE) in Nanyang Technological University, with MMLab@NTU. Previously, he was a research fellow (2018-2020) in CUHK (with Prof. Dahua Lin) and a post-doc researcher (2017-2018) in UC Berkeley (with Prof. Stella X. Yu). His research interests include computer vision, machine learning and computer graphics. Ziwei received his Ph.D. (2013-2017) from CUHK / Multimedia Lab, advised by Prof. Xiaoou Tang and Prof. Xiaogang Wang. He is fortunate to have internships at Microsoft Research and Google Research. His works include Burst Denoising, CelebA, DeepFashion, Fashion Landmarks, DeepMRF, Voxel Flow, Long-Tailed Recognition, and Compound Domain Adaptation. His works have been transferred to products, including Microsoft Pix, SenseFocus, and Google Clips.
\end{IEEEbiography}